\documentclass[runningheads]{llncs}
\usepackage[T1]{fontenc}
\usepackage{graphicx}
\usepackage{hyperref}
\usepackage{color}
\renewcommand\UrlFont{\color{blue}\rmfamily}
\usepackage[inline]{enumitem}
\usepackage[noend]{algpseudocode}
\usepackage[super]{nth}
\usepackage[table]{xcolor}
\usepackage{algorithm}
\usepackage{amsfonts}
\usepackage{amsmath}
\usepackage{amssymb}
\usepackage{booktabs}
\usepackage{csquotes}
\usepackage{mathtools}
\usepackage{multirow}
\usepackage{pgfplots}
\usepackage{tikz}
\usepackage{siunitx}
\usepackage{diagbox}
\usepackage[caption=false]{subfig}
\usepackage{svg}
\usepackage{tabularx}
\usepackage{xcolor}
\usepackage{caption}
\usepackage{wrapfig}
\fussy
\usepgfplotslibrary{groupplots}
\usetikzlibrary{matrix}
\pgfplotsset{compat=1.14}
\let\oldReturn\Return
\renewcommand{\Return}{\State\oldReturn}

\newcommand{\vect}{\mathbf}
\definecolor{darkblue}{HTML}{33629d}
\newrobustcmd{\ColorComment}[1]{{\color{darkblue}\Comment{#1}}}
\makeatletter
\newcommand{\multiline}[1]{\begin{tabularx}{\dimexpr\linewidth-\ALG@thistlm}[t]{@{}X@{}} #1\end{tabularx}}
\makeatother
\begin{document}
\title{BAARD: Blocking Adversarial Examples by Testing for Applicability, Reliability and Decidability}
\titlerunning{BAARD: Blocking Adversarial Examples}
%
\author{
    Xinglong Chang\inst{1} \and
    Katharina Dost\inst{1} \and
    Kaiqi Zhao\inst{1} \and
    Ambra Demontis\inst{2} \and
    Fabio Roli\inst{3} \and
    Gillian Dobbie\inst{1} \and
    J\"org Wicker\inst{1}
}
%
\authorrunning{X. Chang et al.}
\institute{
    The University of Auckland, Auckland, New Zealand \\
    \email{xcha011@aucklanduni.ac.nz, \{katharina.dost, kaiqi.zhao, g.dobbie, j.wicker\}@auckland.ac.nz} \and
    University of Cagliari, Cagliari, Italy \\
    \email{ambra.demontis@unica.it} \and
    University of Genoa, Genoa, Italy \\
    \email{fabio.roli@unige.it}
}
\maketitle
\begin{abstract}
    Adversarial defenses protect machine learning models from adversarial attacks, but are often tailored to one type of model or attack. The lack of information on unknown potential attacks makes detecting adversarial examples challenging. Additionally, attackers do not need to follow the rules made by the defender.  To address this problem, we take inspiration from the concept of Applicability Domain in cheminformatics. Cheminformatics models struggle to make accurate predictions because only a limited number of compounds are known and available for training. Applicability Domain defines a domain based on the known compounds and rejects any unknown compound that falls outside the domain. Similarly, adversarial examples start as harmless inputs, but can be manipulated to evade reliable classification by moving outside the domain of the classifier. We are the first to identify the similarity between Applicability Domain and adversarial detection. Instead of focusing on unknown attacks, we focus on what is known, the training data. We propose a simple yet robust triple-stage data-driven framework that checks the input globally and locally, and confirms that they are coherent with the model's output. This framework can be applied to any classification model and is not limited to specific attacks. We demonstrate these three stages work as one unit, effectively detecting various attacks, even for a white-box scenario.

\end{abstract}

\section{Introduction}
\label{section:Introduction}

Machine learning algorithms have shown promising results in many mission-critical fields, such as virtual drug screening \cite{alvarsson2021predicting} and autonomous driving \cite{kloukiniotis2022countering}.
Unfortunately, despite their high accuracy on benign examples, they are vulnerable to adversarial attacks, where malicious users exploit the classifiers' weakness by manipulating the input data \cite{demontis2019adversarial}.
Starting from a benign data point, attackers craft a small perturbation that allows them to achieve the desired outcome: misclassification of the input example.
For example, by adding a small artifact to a stop sign, a self-driving vehicle can be fooled into misclassifying the stop sign as a speed limit sign, with the risk of causing a car crash \cite{kloukiniotis2022countering}.

Adversarial detectors extract features from unlabeled examples and use them to identify adversarial examples based on certain thresholds \cite{tramer2020adaptive}.
Existing detectors often suffer from the following issues:
First, many detectors focus on detecting adversarial examples with only minimal perturbations \cite{tramer2022detecting} and tend to fail to detect stronger ones.
Second, many defenses are built on a single assumption or one attack, i.e., adversarial examples lead to overly confident predictions from the classifier \cite{hu2019new}.
However, attackers are not constrained by such assumptions, as they can easily bypass such a detector by altering their strategy.
Third, most defenses are tailored to a specific machine learning architecture and do not generalize to other models \cite{yang2020ml}.
There is a lack of flexible detectors that can detect unseen attacks on various classifiers.

In cheminformatics, models are trained on a finite number of compounds because the data-collecting process is expensive and time-consuming.
However, the chemical space is vast and diverse in its properties, so models trained on one part of the space may not work on others.
Hence, models typically struggle to generalize unseen compounds.
To avoid false predictions, {\em Applicability Domain} (AD) is a concept that defines a domain in which a model can perform reliably.
Compounds that are outside this domain are rejected, as the model cannot make reliable predictions on them \cite{alvarsson2021predicting}.
Similar to cheminformatics, adversarial detectors only have the information for known attacks.
However, new attacks come out so frequently that it is impossible to cover all attacks.
In this paper, instead of defending against previously unseen attacks, we focus on what the classifier can reliably predict, the training data.
Inspired by the idea of a triple-stage AD originally introduced by Hanser {\em et al.} \cite{hanser2016applicability}, we propose the \textsc{Baard} framework, \textbf{B}locking \textbf{A}dversarial examples by testing for \textbf{A}pplicability, \textbf{R}eliability and \textbf{D}ecidability.

\begin{figure}[t!]
    \centering
    \includegraphics[width=0.6\columnwidth]{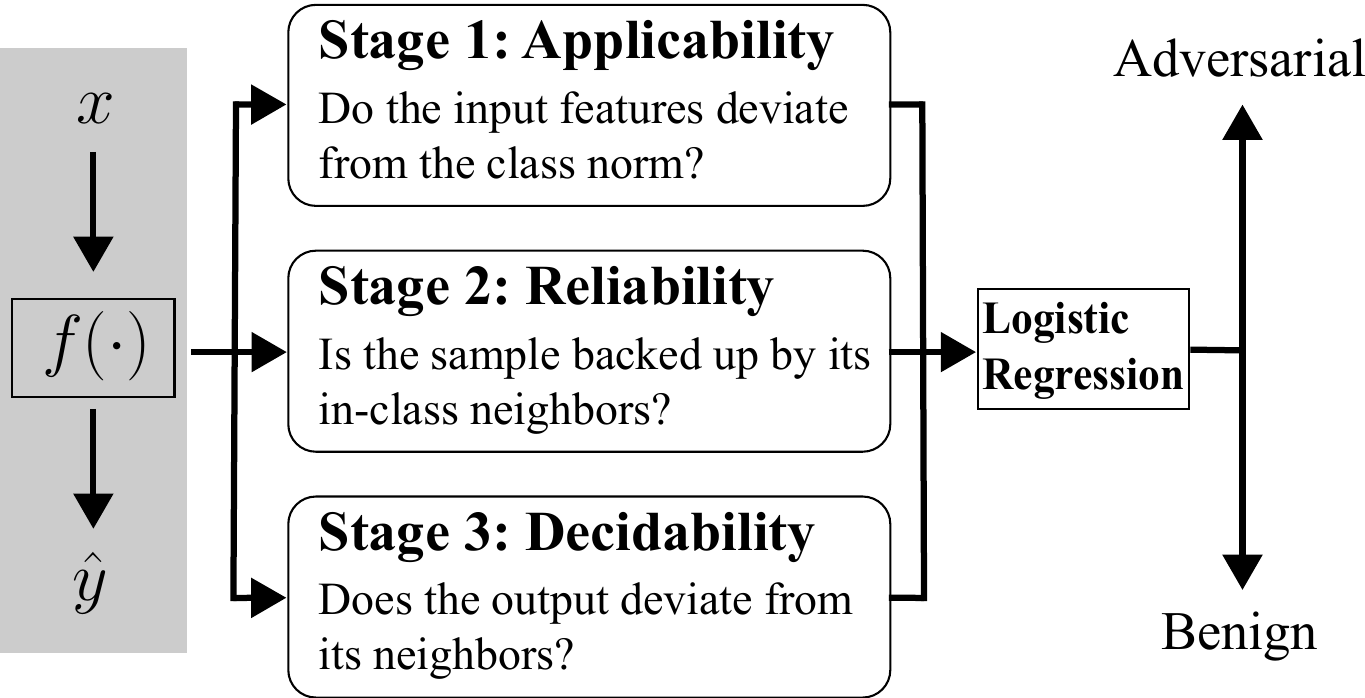}
    \caption{An overview of \textsc{Baard}.
        \textsc{Baard} analyzes an example $\vect{x}$, the classifier $f(\cdot)$, and its prediction $\hat{y}$ together by checking the Applicability, Reliability, and Decidability. Each stage outputs a score. The scores are used to train a logistic regression model to predict whether $\vect{x}$ is benign or adversarial.
    }
    \label{fig:baard_flowchart}
\end{figure}

To identify unknown attacks, \textsc{Baard} investigates the example from three different perspectives, utilizing the training data in the following ways:
\begin{enumerate*}
    \item {\em Applicability Stage} uses the training data to validate the input globally;
    \item {\em Reliability Stage} confirms that the example can be backed up by training data locally; and
    \item {\em Decidability Stage} checks the model's output to ensure it is coherent with the input.
\end{enumerate*}
These three stages work as one unit to inspect the model's interpretation of an unlabeled example.
As shown in Fig.~\ref{fig:baard_flowchart}, \textsc{Baard} rejects the example if there is an inconsistency between the input and the model's prediction.

We summarize our contributions as follows:
\begin{itemize}
    \item We are the first to demonstrate the effectiveness of linking two previously unlinked fields: the Applicability Domain in cheminformatics and adversarial detection in machine learning.
    \item Inspired by the Applicability Domain, we propose the \textsc{Baard} framework (Blocking Adversarial examples by testing Applicability, Reliability, and Decidability), which utilizes training data to systematically detect adversarial examples from three different perspectives.
    \item By designing an adaptive white-box attack targeting \textsc{Baard}, we show that it is difficult to penetrate all three stages, even under the worst scenario.
    \item We demonstrate \textsc{Baard} is highly portable. This simple yet effective framework can detect adversarial examples with various constraints on a wide range of classifiers, including classifiers that have been neglected previously despite being vulnerable to attacks, such as support vector machines and decision trees.
\end{itemize}

We introduce the adversarial threat model, attacks and detectors relevant to this paper in Sec.~\ref{section:Background}.
Sec.~\ref{section:Methodology} and \ref{section:Experiments} present the \textsc{Baard} framework and demonstrate its effectiveness, respectively.
Sec.~\ref{section:Conclusion} concludes this paper.

\section{Background}
\label{section:Background}

This paper focuses on detecting {\em evasion attacks}, where the attacker crafts malicious inputs by adding perturbations to existing examples which can deceive the classifier to make unexpected predictions \cite{demontis2019adversarial}.
Evasion attacks are the most common adversarial attacks since it is easier for a malicious user to interact with the model at inference time.

\subsubsection{Evasion Attacks.}
One of the earliest attacks on {\em neural network} (NN) models is the {\em Fast Gradient Sign Method} (FGSM) \cite{goodfellow2015explaining}, a single-step attack that forms the adversarial example as: $\vect{x}' = \vect{x} - \epsilon \cdot \text{sign}(\bigtriangledown_{\vect{x}}\ell(\vect{x},y))$
where $\vect{x}$ is a benign input, $y$ is the targeted label, $\ell(\vect{x},y)$ is the loss function used by the classifier, and  hyperparameter $\epsilon$ controls the amount of perturbation.
    {\em Auto Projected Gradient Descent} (APGD) \cite{croce2020reliable} is the latest improved version of {\em Projected Gradient Descent} (PGD) \cite{madry2018towards}.
PGD is a multi-step variant of FGSM.
It achieves a higher success rate by iteratively solving the optimization problem.
Improving on PGD, APGD dynamically adjusts the number of iterations to ensure minimal perturbation while maintaining the success rate.
Directly optimizing on the input space can be difficult, since NN models are highly non-linear.
Instead of optimizing on the input space, the {\em Carlini and Wagner Attack} (CW) \cite{carlini2017towards} transforms the image from the pixel space to the simpler $\tanh$ space.
Not only NN models are vulnerable to adversarial attacks, the {\em Decision Tree Attack} (DTA) \cite{papernot2016transferability} exploits the data structure of a decision tree.
The algorithm makes minimal changes at each node and keeps traversing from the leaf to the root until the prediction from the classifier deviates from the legitimate class.

\subsubsection{Detection.}
Detecting adversarial examples with indistinguishable perturbations (hard to recognize by human) has been studied extensively \cite{tramer2022detecting}.
One common assumption is that if the adversarial perturbation is small enough, the legitimate class can be restored by adding or removing noise.
Detectors, such as {\em Feature Squeezing} (FS) \cite{xu2017feature} and the {\em Positive and Negative representation} (PN) detector \cite{luo2022detecting} are motivated by image reconstruction techniques.
FS is a defense motivated by using image filters to restore adversarial examples.
He {\em et al.} \cite{he2017adversarial} pointed out that strong adversaries can easily bypass FS.
The PN detector assumes an adversary cannot simultaneously deceive a classifier trained on both the original and
color-negative images.
Such techniques have clear limitations, a detector that uses images' properties cannot be generalized to other data types.

Another direction is to combine neighborhood relationship and noise generation.
    {\em Region-based Classification} (RC) \cite{cao2017mitigating} replaces the classifier with a region-based classifier by generating noisy samples centered at the example, and a decision is made via majority voting.
Similar to RC, the {\em Odds are odd} (Odds) \cite{roth2019odds} detector assumes that adversarial examples are less robust to noise than benign examples.
The assumption is that latent outputs significantly change when adding noise
to an adversarial example.
    {\em Local Intrinsic Dimensionality} (LID) \cite{ma2018characterizing} is another neighbor-based algorithm that uses the intrinsic dimension metric by combining latent outputs from all hidden layers of a NN.
The statistics are learned by comparing benign, noisy, and adversarial examples.
ML-LOO \cite{yang2020ml} computes Leave-One-Out feature attribution maps on multiple hidden layers of a NN, and uses them to distinguish between benign and adversarial examples.
Many detectors are based on certain assumptions of one type of attack.
If the attacker's goal is to bypass the system, such a constraint may not apply \cite{carlini2017adversarial}.
A detection that is tailored to one attack is not robust against white-box attacks, where the attacker knows a particular defense is placed \cite{tramer2020adaptive}.

\section{\textsc{Baard}: Blocking Adversarial Examples}
\label{section:Methodology}

This paper connects cheminformatics' Applicability Domain with adversarial detection in machine learning.
The goal of AD is to reject chemical compounds that the classifier cannot reliably predict.
Therefore, AD analyzes the feature space and the classifier together to define a tight region around the training instances but omits the rest of the space \cite{netzeva2005current}.
Adversarial examples are perturbations of legitimate example, and remain similar to the original example.
However, adversarial examples are designed to cause misclassifications leading to inconsistencies between the predicted labels of the adversarial example and its legitimate neighbors.
This observation leads us to believe that the idea used in AD can effectively detect adversarial examples.
\textsc{Baard} consists of three stages as shown in Fig.~\ref{fig:baard_flowchart}.
The rest of this section explains the working of each stage and their effectiveness when combined together.

\subsubsection{Applicability Stage.}
In chemistry, this stage checks the compound to confirm it is appropriate for the model to make a prediction \cite{hanser2016applicability}.
Here, we know the model is trained on the training data, so we check the input feature space by comparing it with the training data globally.
We conduct a Z-test by computing mean and standard deviation of input features for each class from the training data.
Given an example $\vect{x}$, the Z-score is defined by
$
    \vect{z}_{\vect{x}, \hat{y}} \coloneqq (x - \mu_{X_{\text{train}}, \hat{y}}) / \sigma_{X_{\text{train}}, \hat{y}}
$,
where $\mu_{X_{\text{train}}, \hat{y}}$ and $\sigma_{X_{\text{train}}, \hat{y}}$ are the mean and standard deviation for examples in the training data that have the same label as the model's prediction $\hat{y}$, and $\vect{z}_{\vect{x}, \hat{y}}$ has the same dimension as $\vect{x}$.
Because we are only interested in the extrema and Z-test is two-tailed, we define the Applicability Score as: $\text{S1 score} \coloneqq \texttt{max}(|\vect{z}_{\vect{x}, \hat{y}}|)$.
The Applicability Stage inspects each feature of the new, unlabeled example, individually. It outputs a high score if any feature is significantly different from the training samples that match the classifier's predicted label.

\subsubsection{Reliability Stage.}
Given a compound, this stage quantifies the relevance of information available to the model in chemistry.
We implement this stage by examining the input locally using the compound's neighbors in the training set.
Unlike the previous stage, which considered each input feature independently, this stage accounts for all features together using the neighborhood relationship.

Adversarial examples aim to minimize the perturbation while forcing the model to make classification errors \cite{goodfellow2015explaining}.
This moves the legitimate input closer to the decision boundary, causing the predicted label to change and potentially placing the example far away from its new in-class neighbors.
The reliability test is based on the distances between adversarial examples and their neighbors.
These distances are often higher than the distances between legitimate examples and their neighbors.

\begin{algorithm}[t!]
    \footnotesize
    \begin{algorithmic}[1]
        \Require
        $\vect{x}$: unlabeled example,
        $\hat{y}$: its prediction,
        $(X, Y)$: training set,
        $k_{\text{S2}}$: number of neighbors, and
        $m_{\text{S2}}$: sample size.
        \Ensure
        \texttt{S2\_score} $\in [0, 2\pi]$

        \State $X_{\hat{y}} \leftarrow \text{Random sampling } \{ (\vect{x}_1, y_1), \ldots, (\vect{x}_{m_{\text{S2}}}, y_{m_{\text{S2}}}) \}$, where $\vect{x}_i \in X \text{, } y_i \in Y \text{, and } y_i = \hat{y}$
        \State $D(\vect{x}, X_{\hat{y}}) \leftarrow$ Compute angular distances between example $\vect{x}$ and subset $X_{\hat{y}}$
        \State $\texttt{S2\_score} \leftarrow \texttt{mean}(\texttt{top\_k}(D(\vect{x}, X_{\hat{y}}), k_{\text{S2}}))$ \ColorComment{Compute the mean of top $k_{\text{S2}}$ distances}
        \Return \texttt{S2\_score}
    \end{algorithmic}
    \caption{\textsc{Baard} Stage 2 -- Reliability Stage}
    \label{alg:reliability}
\end{algorithm}

Choosing an appropriate distance metric is essential when measuring nearest neighbors.
The Euclidean distance ($L_2$-norm) is well suited for low-dimensional space, but
Cosine similarity has shown more robust results in high-dimensional sparse features \cite{luo2022detecting}.
Cosine similarity between two feature vectors $A$ and $B$ is defined as:
$
    S_C(A, B) \coloneqq \sum^{n}_{i=1}A_i B_i / [ (\sum^{n}_{i=1}A^2_i)^{\frac{1}{2}} (\sum^{n}_{i=1}B^2_i)^{\frac{1}{2}} ]
$,
where $n$ is the dimension of the feature vector, and $S_C \in [-1, 1]$.
If $S_C$ is close to 1, $A$ and $B$ are positive co-linear vectors.
If $S_C = 0$, they are independent vectors,
and if $S_C \approx -1$, they are strong opposite vectors.
This means neither minimal nor maximal indicates $A$ and $B$ are close.
To properly present the distance between two features using cosine similarity, we compute the angular distance $D$, which is defined as:
$
    D(A, B) \coloneqq \texttt{arccos}(S_C(A, B)) / \pi
$.

Algorithm~\ref{alg:reliability} provides the pseudocode for this stage.
It takes two hyperparameters:
the number of nearest neighbors $k \in \mathbb{N}$, and the sample size $m$ that limits the computational expense.
For an unlabeled example $\vect{x}$, the S2 score is the mean distance of the $k$-nearest neighbors of $\vect{x}$ within a subset of training data where examples have the same label as the prediction $\hat{y}$.
Because the angular distance is within $[0, 2\pi]$, the S2 score shares a similar scale as the S1 score.
To reduce the computational cost, we randomly sample $m$ instances from the training examples where the legitimate labels are the same as $\hat{y}$.

\subsubsection{Decidability Stage.}
This stage confirms whether the model's output is coherent with the evidence from previous stages.
Machine learning models operate under the assumption that similar examples have similar labels.
Hence, a trained model can generalize to new and previously unseen examples.
However, this is often violated when the model tries to predict maliciously crafted adversarial examples.
The prediction of an adversarial example often conflicts with the predictions of its neighbors.
As shown in Fig.~\ref{fig:baard_flowchart}, we use the local neighborhood relationship to check adversarial examples based on this property.

Algorithm~\ref{alg:decidability} uses the same distance metric as in previous stages.
The critical difference is that the entire training data are used regardless of their labels.
We apply the Softmax function so the model outputs probability estimates.
Given an example, we run a Z-test on its probability estimates based on its $k$ neighbors.

\begin{algorithm}[t!]
    \footnotesize
    \begin{algorithmic}[1]
        \Require
        $\vect{x}$: unlabeled example,
        $\hat{y}$: its prediction,
        $(X, Y)$: training set,
        $k_{\text{S3}}$: number of neighbors, and
        $m_{\text{S3}}$: sample size.
        \Ensure
        \texttt{S3\_score}

        \State $S \leftarrow \text{Random sampling } \{\vect{x}_1, \ldots, \vect{x}_{m_{\text{S3}}} \} \text{ where } \vect{x}_i \in X$
        \State $D(\vect{x}, S) \leftarrow$ Compute angular distances between example $\vect{x}$ and subset $S$
        \State $X' \leftarrow \texttt{top\_k}(D(\vect{x}, S), k_{\text{S3}})$ \ColorComment{Find top $k$-nearest neighbors.}
        \State $P' \leftarrow \texttt{Softmax}(f(X'))$ \ColorComment{Compute probability estimates for neighbors.}
        \State $\mu_{P'}, \sigma_{P'} \leftarrow \texttt{mean}(P'), \texttt{std}(P')$ \ColorComment{Compute mean and standard deviation vectors.}
        \State $\vect{z} \leftarrow |\frac{\texttt{Softmax}(f(\vect{x})) - \mu_{P'}}{\sigma_{P'}}|$
        \Return $\vect{z}_{\hat{y}}$ \ColorComment{$\vect{z}_{\hat{y}}$ is the value of $\vect{z}$ index at $\hat{y}$.}
    \end{algorithmic}
    \caption{\textsc{Baard} Stage 3 -- Decidability Stage}
    \label{alg:decidability}
\end{algorithm}

\subsubsection{Combining All Stages.}
A single stage may be effective on a certain type of attack, but no stage alone can cover all attacks.
The Applicability and Reliability Stages both check the feature space but from different perspectives.
Once we collect enough evidence from the input space, the Decidability Stage checks the output to ensure the model's output is coherent with the evidence.
We fit a Logistic Regression model using the scores from \textsc{Baard} on a hold-out training set to distinguish adversarial examples from legitimate inputs.

While being fast and memory-efficient, this approach has two issues when dealing with image data.
\begin{enumerate*}
    \item When the feature space is sparse, the S1 score becomes noise-sensitive.
    \item The score varies under transformations, such as translation and rotation.
\end{enumerate*}
Images are commonly modeled by convolutional neural networks, because the convolutional layers can learn internal representation in a two-dimensional space.
Hence, these latent outputs represent the extracted feature space learned by the model.
We overcome the above issues by using the latent outputs after the convolutional layers but before the fully connected layer.
Note that tabular data does not suffer from the same issues.
Moreover, anything related to the training data can be calculated beforehand to speed up the algorithm at  inference time.

\section{Experiments}
\label{section:Experiments}

We evaluate \textsc{Baard} by analyzing its parameters, deconstructing it, and testing it against attacks in both white-box and gray-box settings.
We repeat the experiments five times to ensure robustness.
To ensure reproducibility, all data, pre-trained classifiers, hyperparameters, additional results, and code are available at \href{https://github.com/changx03/baard}{\UrlFont{https://github.com/changx03/baard}}.

\subsection{Experimental Setup}
\label{subsection:ExperimentSetup}

\subsubsection{Data and Classifiers.}
We test \textsc{Baard} on both image and tabular data.
We acquire MNIST and CIFAR10 with default train-test split from PyTorch for image datasets.
We use the model from Carlini and Wagner \cite{carlini2017towards} for MNIST and ResNet18 from PyTorch for CIFAR10.
The pre-trained models are available in our repository.
We remove the misclassified examples and sample 1000 images for generating adversarial examples and another 1000 for validating the detectors from the test set.
We acquire all tabular data from the UCI ML repository
\footnote{Source: \href{https://archive.ics.uci.edu/ml/index.php}{\UrlFont{https://archive.ics.uci.edu/ml}}}.
All tabular data use a 60-20-20 split.
The SVM and {\em Decision Trees} (DT) models for tabular data use the default parameters.
Additional datasets are tested and included in our repository.

\subsubsection{Attack Algorithms.}
We evaluated \textsc{Baard} and other detectors under various attacks that are covered in Sec.~\ref{section:Background}, including PGD \cite{madry2018towards}, APGD \cite{croce2020reliable}, CW-$L_2$ \cite{carlini2017towards}, and DTA \cite{papernot2016transferability}.
We additionally include the results for FGSM \cite{goodfellow2015explaining}, Boundary Attack \cite{brendel2018decision}, and DeepFool \cite{moosavi2016deepfool} in our repository.
We define the adversary's goal to have examples misclassified as any class except the true one so all attacks are untargeted.
To test attack strengths, when there are multiple L-norm constraints, we test both $L_\infty$ and $L_2$ norm constraints.
For each attack, we have considered a wide range of attack strengths.
For instance, the parameter $\epsilon$ in APGD controls the amount of perturbation allowed \cite{croce2020reliable}.
We set the minimal value to where the attack has at least $95\%$ success rate.
The minimal $\epsilon$ for APGD is set to $0.22$ and $4.0$ for $L_\infty$ and $L_2$ on MNIST, $0.01$ and $0.3$ on CIFAR10, respectively.
In Table~\ref{tab:gray-box-table}, these values are used as the ``Low'' $\epsilon$, and the ``High'' is set to at least double the ``Low'' where there is a visible artifact on the example, but the legitimate label is still recognizable.

\subsubsection{Evaluation Metrics.}
We report the {\em Area Under the Curve} (AUC) of the {\em Receiver Operating Characteristic} (ROC) curve as the performance metric.
In practice, a single threshold may be selected based on the {\em False Positive Rate} (FPR).
Hence, we also report TPRs when thresholds are chosen based on $5\%$ FPRs (TPR@5FPR) when comparing different detectors.

\subsection{Detection Results}
\label{subsection:DetectionResults}

\subsubsection{Parameter Analysis.}
We treat each stage as an individual detector when tuning the hyperparameters.
Since each stage's performance directly links to $k$ and the sample size $m$ is for speeding up the algorithm, we first find the optimal $k$ while using the entire training set, then use the optimal $k$ to tune $m$.

\begin{figure}[t!]
    \centering
    \includegraphics[width=\columnwidth]{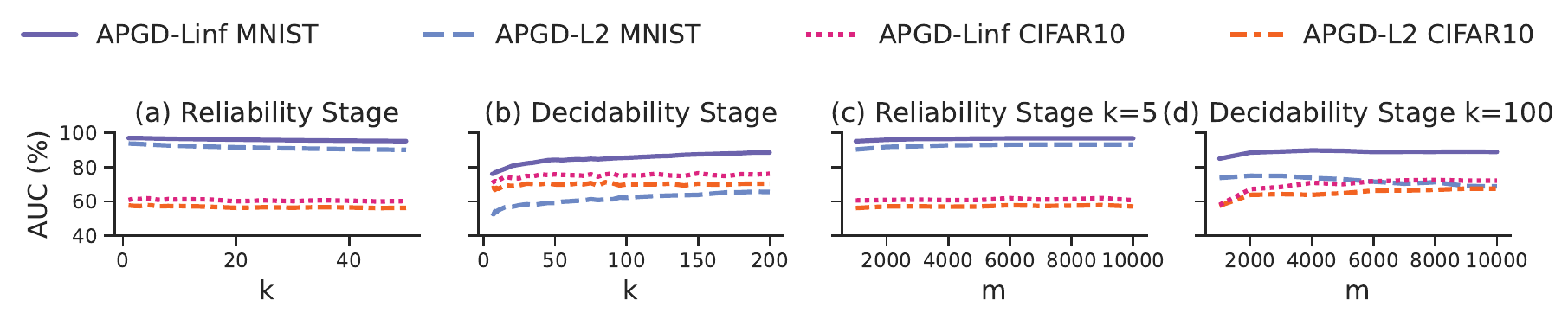}
    \caption{Tuning hyperparameters for \textsc{Baard} at the minimal adversarial perturbation. We first search for the optimal $k$, then tune the sample size $m$.}
    \label{fig:param}
\end{figure}

The values of $k_{\text{S2}}$ and $k_{\text{S3}}$ are different.
As shown in Fig.~\ref{fig:param}, $k_{\text{S2}}$ in the Reliability Stage becomes stable after the initial fluctuation.
Reliability prefers a smaller $k_{\text{S2}}$ value, as it checks the closest representation of $\vect{x}$ in the training samples with the same label as $\hat{y}$.
Because Decidability finds neighbors from all training samples, a greater value of $k_{\text{S3}}$ is preferred.
Once $k_{\text{S2}}$ and $k_{\text{S3}}$ are chosen, the optimal $m_{\text{S2}}$ and $m_{\text{S3}}$ should be the minimum value while maintaining the detector's performance.
Because the Reliability Stage uses the in-class training subset, the possible sample size is smaller than the Decidability Stage.
Our results show that the detector's performance is sturdy after initial turbulence, suggesting that the sub-sampling has minimal impact on the overall performance.
The experiment concludes that \textsc{Baard} requires minimal tuning.
We set $k$ to 5 and 100 and $m$ to 1000 and 5000 for the Reliability and Decidability Stages respectively for all image datasets.

\subsubsection{Ablation Study.}
\begin{figure}[t!]
    \centering
    \includegraphics[width=\columnwidth]{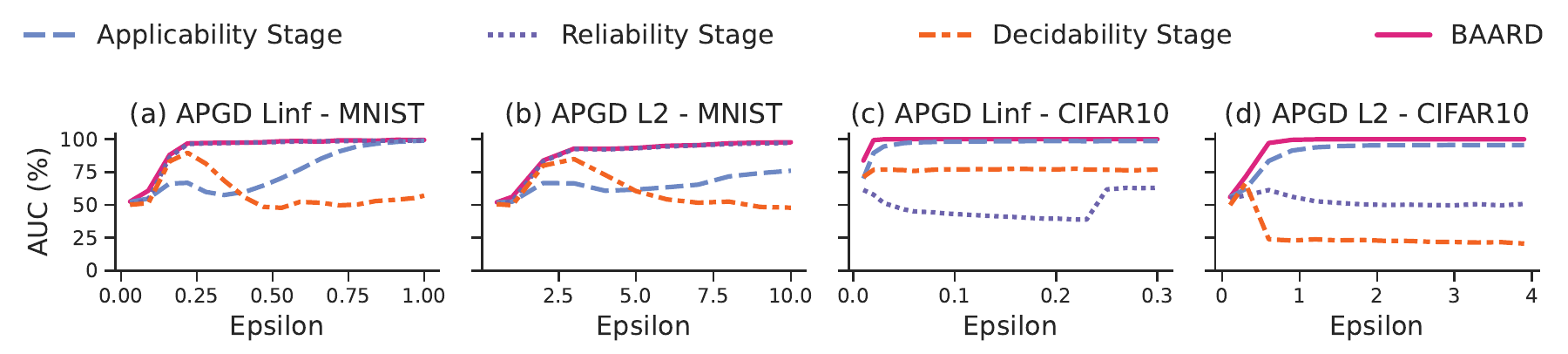}
    \caption{\textsc{Baard}'s performance under decomposition against adversarial attacks with a full range of perturbations.}
    \label{fig:ablation}
\end{figure}

We decompose \textsc{Baard} to investigate how each stage contributes to the overall performance.
Fig.~\ref{fig:ablation} shows AUCs at various adversarial perturbations.
Since attacks under a $L_{\infty}$ constraint result in a significant deviation on the feature space \cite{yang2020ml}, we find neither the S1 nor S2 score alone can detect such attacks.
In Fig.~\ref{fig:ablation}d, the Decidability Stage's AUC (orange dotted line) goes lower than $50\%$ when $\epsilon \ge 0.6$,
indicating that the correlation between the S3 score and the detector's performance flip when $\epsilon$ increases.
It means the classifier becomes more confident with the misclassified predictions when $\epsilon$ increases, leading to smaller S3 scores.
Meanwhile, the S1 score becomes larger since the attack makes significant changes to the input.
A low AUC on one stage indicates that stage alone is insufficient as a detector.
However, by combining all stages, the results show \textsc{Baard} is effective on a wide range of adversarial perturbations.

\subsubsection{White-Box Evaluation.}
\begin{wrapfigure}{r}{0.45\columnwidth}
    \centering
    \vspace{-1em}
    \includegraphics[width=0.45\columnwidth]{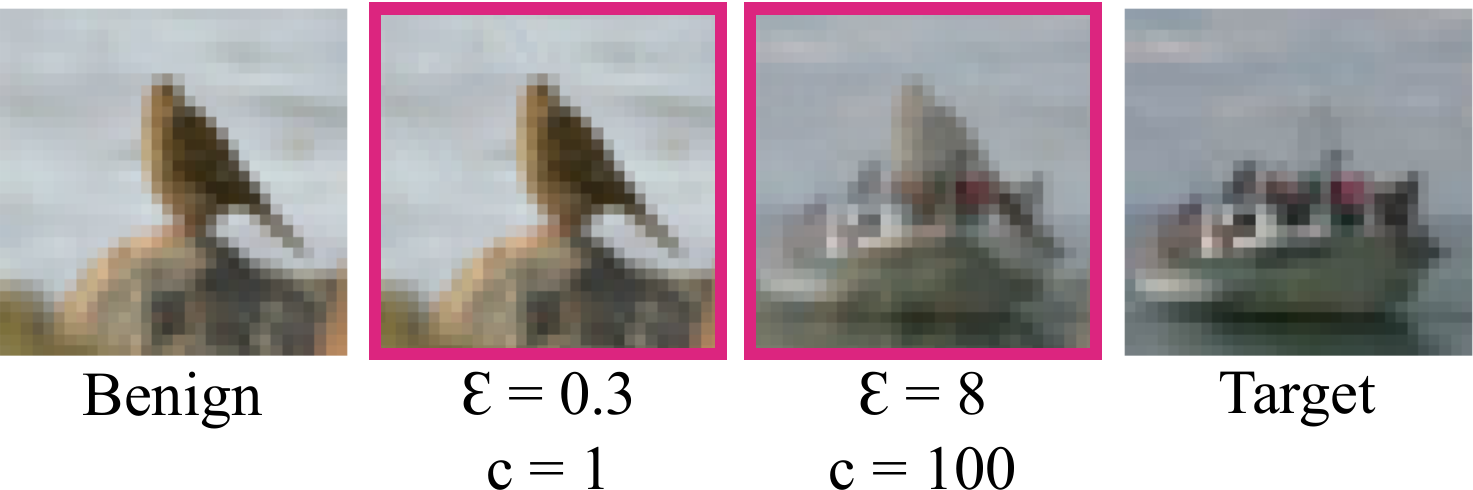}
    \caption{Apply our Adaptive White-box Targeted $L_2$ Attack to CIFAR10; When extreme parameters are used, it transforms a benign example into the target.}
    \vspace{-1em}
    \label{fig:whiteboxExample}
\end{wrapfigure}
We address the robustness of \textsc{Baard} against adaptive white-box attacks.
To simultaneously attack the classifier and \textsc{Baard}, the attacks' loss function is $\mathcal{L}^{*} \coloneqq \mathcal{L} + \mathcal{L}_{\text{S1}} + \mathcal{L}_{\text{S2}} + \mathcal{L}_{\text{S3}}$, where $\mathcal{L}$ is the term for the evasion attack: $\mathcal{L} \coloneqq -\text{CrossEntropy}(f(\vect{x}'), y_{\text{target}})$, and the rest of the terms are the losses for each stage.
Because none of the stages are differentiable, a common approach is to apply gradient approximation \cite{hu2019new}.
Tramer {\em et al.} \cite{tramer2020adaptive} pointed out that gradient approximation tends to fail when the loss function includes multiple indifferentiable terms, a more robust approach is to find a target $\vect{x}_{\text{target}}$ that can pass the detector and use it as a reference.
Hence, we propose an {\em Adaptive White-box Targeted} (AWT) attack as follows: we find the nearest neighbor from the training data based on the same feature space \textsc{Baard} uses, as $\vect{x}_{\text{target}}$, and then minimize the difference between $\vect{x}$ and $\vect{x}_{\text{target}}$ to bypass S1 and S2.
To avoid $f(\cdot)$ making over confident predictions, we use $f(\vect{x}_{\text{target}})$ as a reference to bypass S3.
The new loss function becomes
$\mathcal{L}^{*} \coloneqq - \ell(f(\vect{x}), f(\vect{x}_{\text{target}})) - c \ell(\vect{x}, \vect{x}_{\text{target}})
$, where both terms use the {\em Mean Squared Error} (MSE) loss and the hyperparameter $c$ controls the ratio on how much $\vect{x}$ moves toward to $\vect{x}_{\text{target}}$.
As shown in Fig.~\ref{fig:whiteboxExample}, if we relax the perturbation constraint $\epsilon$ and dial $c$ to an extreme, the adversarial example becomes indistinguishable from the target.

\begin{figure}[t!]
    \centering
    \includegraphics[width=\columnwidth]{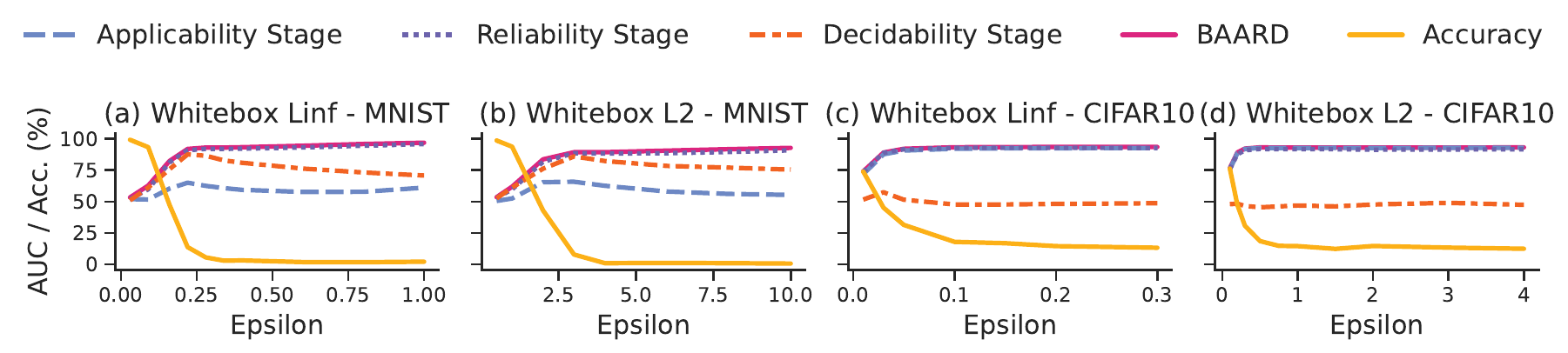}
    \caption{\textsc{Baard}'s performance against Adaptive White-box Targeted attacks. The accuracy indicates the classifier's performance under such attacks.}
    \label{fig:whitebox}
\end{figure}

We present the evaluation of \textsc{Baard} against our AWT attacks in Fig.~\ref{fig:whitebox} with $c$ set to 1.
The attack can successfully deceive the classifier and bypass S1 or S3, but not all stages.
Previous works show similar algorithms are effective on detectors with multiple loss functions, such as the Odds detector \cite{tramer2020adaptive}.
We find such attacks are ineffective on \textsc{Baard}, as three stages work together, which are robust against AWT attacks under both $L_2$ and $L_{\infty}$ constraints.

\begin{table}[t!]
    \centering
    \caption{Performance of detectors. The AUC scores ($\%$) on the left are computed from logistic regression. The right side shows the corresponding TPR at $5\%$ FPR. ``Low'' and ``High'' indicate perturbations allowed for the attack.}
    \includegraphics[width=\columnwidth]{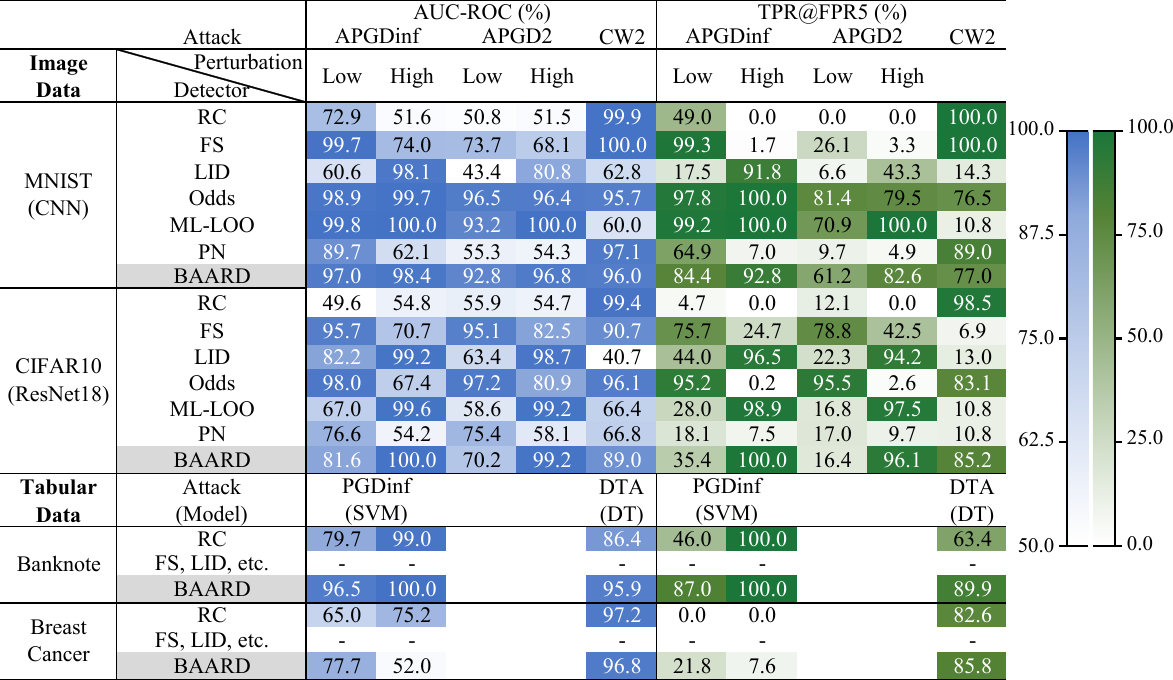}
    \label{tab:gray-box-table}
\end{table}

\subsubsection{Gray-Box Benchmark.}
To benchmark the performance of \textsc{Baard} against other detectors in Sec.~\ref{section:Background}, we use the same hold-out set to train logistic regression models for each dataset based on the features extracted from the detector.
Table~\ref{tab:gray-box-table} presents both the AUC and TPR values obtained by varying the threshold of the regressors' outputs.
\textsc{Baard} performs consistently well across different classifiers under attacks with various strengths, showing outstanding performance on attacks with high perturbations.
One outlier is the APGD attack with an $L_2$ constraint at a low $\epsilon$ on CIFAR10, where most detectors are weak, except FS and Odds.
However, FS and Odds are tuned explicitly for low perturbations and completely fail to detect attacks with high perturbations.
RC can apply to any classifier in theory, but it only performs well in CW2. Meanwhile, the detectors tailored to images and neural networks cannot apply to SVM and DT classifiers.
No detector performs reliably on the PGD attacks on the Breast Cancer dataset.
However, \textsc{Baard} is substantially faster than detectors with similar performance, such as LID, Odds, and ML-LOO.
In conclusion, \textsc{Baard} is the most versatile detector tested that can reliably detect adversarial examples with various constraints on a wide range of classifiers.

\section{Conclusion and Future Work}
\label{section:Conclusion}

In this paper, we connected two previously unlinked domains: the Applicability Domain (AD) in cheminformatics and adversarial detection in machine learning.
By sharing solutions to similar problems, both areas can benefit.
We proposed \textsc{Baard}, a novel adversarial detection framework inspired by AD.
Our experiments showed its robustness against various adversarial evasion attacks, including those with strong perturbations.
\textsc{Baard} is portable and versatile enough to work with any classifier, removing the need for redesigning a defense.
Our framework overcomes challenging issues in the field while maintaining comparable performance.
In future research, we will explore how the insights we have gained from adversarial detection can be transferred into cheminformatics.

\subsubsection{Acknowledgements.}
The authors wish to acknowledge the use of New Zealand eScience Infrastructure (NeSI) national facilities - \href{https://www.nesi.org.nz}{\UrlFont{https://www.nesi.org.nz}}.

{
\bibliographystyle{splncs04}
\bibliography{main.bib}   
}

\end{document}